\documentclass[a4paper,fleqn]{cas-dc}

\usepackage[numbers]{natbib}
\usepackage{graphicx}
\usepackage{amsmath,amssymb}
\usepackage{booktabs}
\usepackage{multirow}
\usepackage{makecell}
\usepackage{tabularx}
\usepackage{xcolor}
\usepackage{url}
\usepackage{hyperref}
\usepackage{subcaption}
\usepackage[utf8]{inputenc}
\usepackage{float}
\usepackage{braket}
\usepackage{array}

\begin{document}

\shorttitle{QWRF-Net for short-term precipitation nowcasting}
\shortauthors{Wang et al.}

\title[mode=title]{QWRF-Net: A Quantum--Wavelet Framework with Rectified Flow for Short-Term Precipitation Nowcasting}

\author[1,2]{Zhuo Wang}
\ead{240503wj@stu.cqut.edu.cn}

\author[1]{Chaorong Li}[orcid=0000-0001-8336-2661]
\cormark[1]
\ead{lichaorong88@163.com}

\author[2]{Wenjie Luo}
\ead{lwj018@stu.cqut.edu.cn}

\author[2]{Chuanhu Deng}
\ead{onlyiyou@stu.cqut.edu.cn}

\cortext[1]{Corresponding author.}

\fntext[1]{This work was supported in part by the Major Project of Yibin University under Grant 2025XJZD01, and in part by the Science and Technology Program of the Yibin Municipal Science and Technology Bureau under Grant 2025JC008.}

\affiliation[1]{organization={School of Computer Science and Technology (School of Artificial Intelligence), Yibin University},
  addressline={},
  city={Yibin},
  postcode={644000},
  state={Sichuan},
  country={China}}

\affiliation[2]{organization={College of Computer Science and Engineering, Chongqing University of Technology},
  addressline={},
  city={Chongqing},
  postcode={400054},
  country={China}}

\begin{abstract}
Short-term precipitation nowcasting is important for hydrometeorological early warning, especially when intense convective rainfall may trigger urban flooding, flash floods, and other high-impact hazards. A key challenge in warning-oriented nowcasting is that radar precipitation fields contain strongly coupled multi-scale structures, while forecast quality often degrades at later lead times, making it difficult to preserve intense precipitation cores and their spatial organization over the full warning-relevant horizon. To address this problem, we propose QWRF-Net, a quantum--wavelet framework with rectified flow for short-term precipitation nowcasting. The core idea is to improve the conditional representation of precipitation by explicitly decomposing latent features into wavelet sub-bands and then performing differentiated quantum-inspired modulation in the decomposed latent space, before generating future sequences through a rectified-flow-based non-autoregressive decoder. Experiments on the KNMI radar and SEVIR benchmarks under a unified evaluation protocol show that QWRF-Net achieves favorable overall performance, with relatively consistent gains at medium-to-high precipitation thresholds, on an extreme-event subset, and in preserving intense precipitation cores and fine-scale structures. Ablation results further indicate that wavelet-based scale disentanglement, differentiated sub-band modulation, and flow-based generation provide complementary benefits within the proposed framework. Overall, these results suggest that jointly enhancing multi-scale precipitation representation and stable multi-step generation is a promising direction for warning-oriented short-term precipitation nowcasting. The observed improvements may also provide a more useful precipitation basis for downstream hydrological and warning-related applications.
\end{abstract}


\maketitle

\section{Introduction}

\subsection{Background and challenges}

In the context of global climate change, the increasing frequency and intensity of extreme weather events continue to pose substantial risks to human society \cite{seneviratne2021weather,shi2017deep}. In particular, short-term intense precipitation triggered by local convection is closely associated with urban flooding, flash floods, landslides, and other high-impact hydrometeorological hazards. Because such events often develop rapidly and evolve within a limited time window, effective early-warning systems require timely and reliable precipitation forecasts \cite{fowler2021anthropogenic}. Consequently, accurate short-term precipitation nowcasting has become an important component of hazard-oriented weather warning and short-term risk mitigation. For flash-flood and urban inundation warning in particular, forecast value depends not only on whether rainfall occurs, but also on whether intense precipitation cores are correctly located and maintained over the warning-relevant lead time. Such properties may also be relevant when nowcasts are used as upstream precipitation inputs for downstream hydrological assessment and warning-related analysis.

From both scientific and operational perspectives, precipitation nowcasting remains challenging. Although numerical weather prediction (NWP) models \cite{al2010review} have achieved considerable success in synoptic- and mesoscale forecasting, their application to rapid convective-scale nowcasting is often constrained by the complexity of physical modeling and the high computational cost of data assimilation \cite{smith2025nowcasting}. Strongly nonlinear convective processes may initiate, intensify, merge, or dissipate within minutes, making them difficult to represent accurately in time-critical forecasting scenarios \cite{amini2022adaptive}. In warning-oriented settings, this challenge is especially important because localized errors in short lead-time rainfall prediction may affect the timing and usefulness of flood-related response. This issue is particularly critical in short-fuse warning situations, where errors in the 30--60 min range may directly reduce the effective preparation window for emergency response.

Deep learning provides a practical data-driven route for precipitation nowcasting \cite{han2023key} by learning spatiotemporal evolution patterns directly from large radar archives. Compared with computationally intensive physics-based forecasting, such models can offer much faster inference in short lead-time settings. However, current data-driven nowcasting methods still face two closely related challenges \cite{prudden2020review}. First, radar precipitation fields contain intertwined multi-scale structures, including broad precipitation organization, localized intense cores, and fine-scale boundaries, which are not always easy to represent adequately within a single feature space. From a warning-oriented hydrometeorological perspective, these structures do not play identical roles: broad precipitation organization is relevant to the overall spatial extent of rainfall, whereas localized intense cores and sharp gradients are more closely associated with short-duration high-impact events. From a hydrological perspective, such large-scale organization provides the rainfall background for runoff generation, whereas localized intense cores more directly influence where short-duration flood-triggering rainfall is concentrated. Second, multi-step forecasting remains vulnerable to progressive degradation, where small prediction errors may accumulate over time and lead to blurred structures, weakened intensities, and reduced stability at later lead times \cite{cheng2025enhanced}. For short-term warning, this issue is particularly important because degraded later lead-time forecasts may reduce the effective response window available for emergency decision-making.

These two difficulties are closely coupled in practice: insufficient conditional representation makes future-sequence generation more difficult, while unstable generation may further obscure precipitation structures that are critical to short-term warning. This motivates a nowcasting framework designed to jointly improve multi-scale precipitation representation and future-sequence generation \cite{zeng2025review}.

\subsection{Research lineage}

Existing precipitation nowcasting methods have evolved from motion extrapolation \cite{sokol2017probabilistic} to discriminative prediction \cite{li2024precipitation} and, more recently, to generative forecasting \cite{asperti2025precipitation}. This progression reflects a gradual shift from modeling precipitation displacement alone to modeling both precipitation structure and future evolution \cite{ye2026improving}.

Early nowcasting methods mainly relied on motion extrapolation, such as optical-flow-based approaches \cite{ha2024deep} and semi-Lagrangian schemes \cite{berenguer2011sbmcast}. These methods can be effective when precipitation evolves smoothly, but they are often less suitable for convective systems involving rapid deformation, growth, merging, and dissipation. In warning-oriented applications, this limitation is important because the rapid emergence or decay of localized convective cores may strongly affect short-term hazard estimates. As a result, purely extrapolative methods may be less reliable when warning decisions depend on the rapid emergence, displacement, or dissipation of localized high-intensity rainfall.

This limitation motivated the development of deep discriminative models that learn future precipitation directly from historical observations. Representative examples include convolutional encoder--decoder models, recurrent nowcasting models, and more recent spatiotemporal prediction architectures \cite{li2024flood,chen2022flood,ngan2023hybrid,liao2024short,mihoc2023convsnow,wang2024spatiotemporal}. While these models have substantially advanced precipitation nowcasting, forecasts trained with point-estimation objectives may smooth intense echoes and fine-scale morphology, and recursive multi-step prediction can still amplify small errors over time \cite{niu2026data}. For warning-oriented hydrometeorological forecasting, such smoothing may reduce fidelity in the location and structure of intense rainfall cores. This loss of structural fidelity is particularly problematic for warning-oriented applications because small spatial shifts in intense cores may lead to large differences in local hazard relevance.

To alleviate the limitations of deterministic prediction, recent studies have increasingly explored generative models \cite{ravuri2021skilful} that learn a distribution over future precipitation states rather than a single point estimate. GAN-based methods such as DGMR \cite{chirigati2021accurate} and more recent approaches such as NowcastNet \cite{zhang2023skilful} indicate that generative formulations are well suited to modeling complex precipitation evolution. Diffusion-based methods such as PreDiff \cite{gao2023prediff}, DiffCast \cite{yu2024diffcast}, and related conditional diffusion models \cite{shi2024codicast} further show strong generation quality, although their iterative denoising procedures may still introduce non-negligible inference cost in time-sensitive settings. In operational hydrometeorological warning, such inference cost may become a practical limitation when frequent forecast updates are required. Flow-based formulations provide an alternative by learning a continuous velocity field and generating future states through ODE-based sampling \cite{richter2026generative}. These developments suggest that improving future-sequence generation is important, but generation quality also depends on how precipitation structures are organized before decoding.

\begin{figure*}[t]
    \centering
    \includegraphics[scale=0.6]{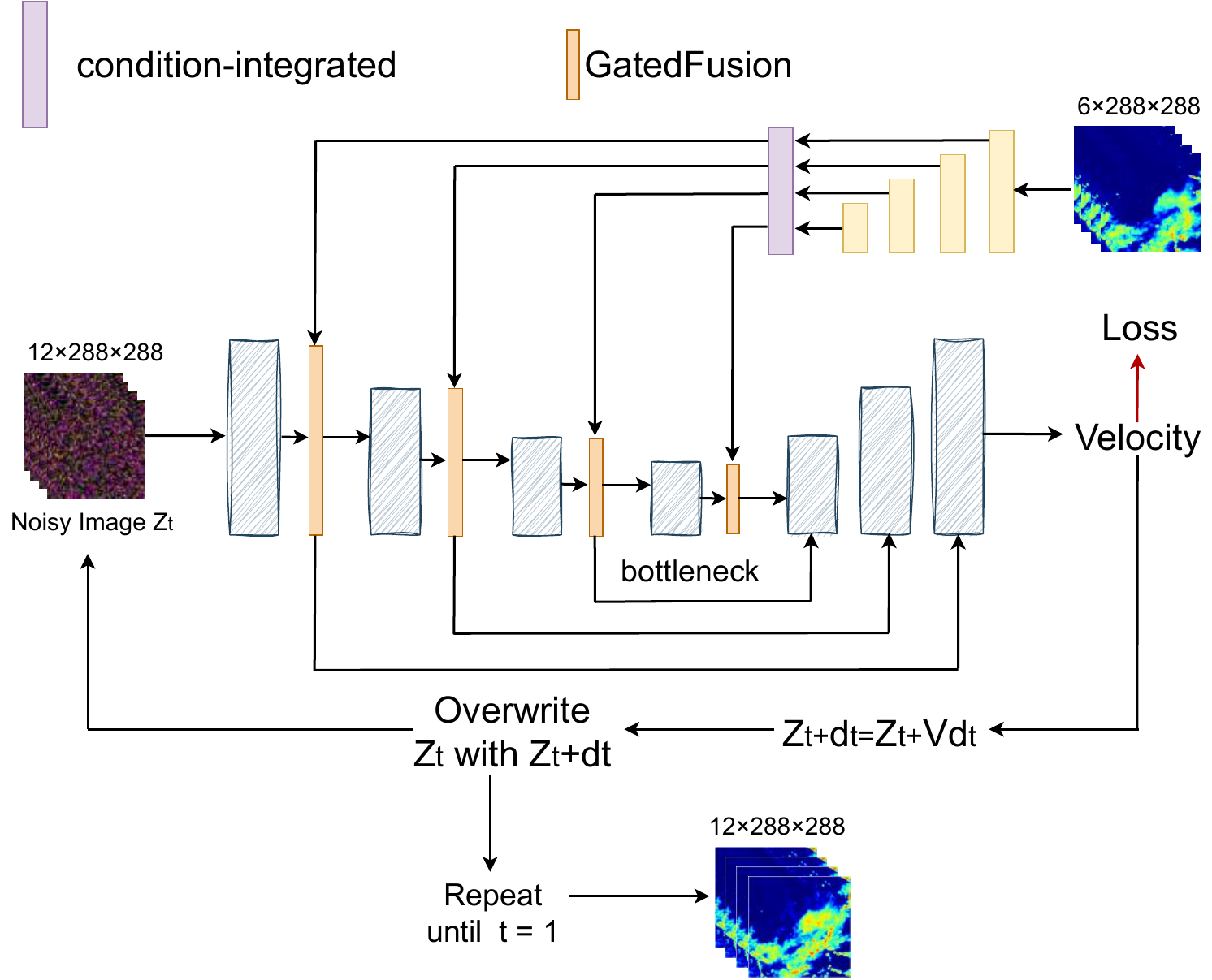}
    \caption{Overall architecture of QWRF-Net for radar-based nowcasting. Given the past 6 frames (30 minutes) as condition, the model generates the next 12 frames (60 minutes). The encoder extracts multi-scale features, the quantum--wavelet bottleneck refines scale-aware precipitation representation, and the flow-based decoder generates the future sequence through ODE-based sampling.}
    \label{fig:overall_arch}
\end{figure*}

\subsection{Our proposed method: QWRF-Net}

To address the above issues, we propose QWRF-Net, a conditional generative framework for short-term radar-based precipitation nowcasting. As illustrated in Fig.~\ref{fig:overall_arch}, QWRF-Net combines a wavelet-decomposed bottleneck with a flow-based decoder to improve scale-aware conditional representation and multi-step future prediction.

From a hydrometeorological forecasting perspective, the central idea is to first organize precipitation information in a way that better reflects its multi-scale warning relevance, and then generate future sequences using a mechanism that is less vulnerable to lead-time degradation. The framework therefore follows a representation-to-generation strategy: it first separates precipitation information by scale, then enhances decomposed components in a differentiated manner, and finally generates future sequences through a non-autoregressive flow-based decoder. In this sense, the proposed framework is designed not only to improve average prediction accuracy, but also to better preserve those rainfall structures that are most consequential for short-term hydrometeorological warning and potentially relevant to downstream hydrological use.

Specifically, a discrete wavelet transform (DWT) \cite{othman2020applications} is introduced at the bottleneck to reorganize mixed latent features into frequency-specific sub-bands, thereby separating low-frequency background information from high-frequency structural details. This step converts an entangled latent representation into structurally differentiated components and provides an explicit basis for distinguishing broad precipitation organization from local warning-relevant sharp structures. After this scale disentanglement, each sub-band is modulated by a learnable quantum-inspired transformation, simulated on classical hardware, so that nonlinear enhancement is performed in a frequency-aware latent space rather than on the original mixed feature map. Here, the quantum-inspired module is used as a classically simulated structured nonlinear operator for sub-band-specific modulation rather than as evidence of practical quantum advantage. Based on the resulting conditional representation, a rectified-flow-based formulation \cite{liu2022rectified} is adopted to learn a conditional velocity field for non-autoregressive future-sequence generation. During inference, the model predicts the target sequence through ODE-based integration rather than recursive frame-by-frame rollout, which is expected to reduce accumulated degradation across the forecast horizon.

Overall, the proposed framework is intended to jointly improve the representation of historical precipitation structures and the generation of future precipitation sequences over multiple lead times. In this way, wavelet decomposition, differentiated sub-band modulation, and flow-based generation serve as sequentially connected components for improving both structural preservation and forecast stability in warning-relevant short-term nowcasting scenarios \cite{liu2024enhanced}.

\subsection{Contributions}

The main contributions of this study are summarized as follows:
\begin{enumerate}
    \item We formulate short-term precipitation nowcasting as a joint problem of conditional precipitation representation and future-sequence generation, with emphasis on preserving multi-scale structures that are relevant to warning-oriented hydrometeorological forecasting.
    \item We introduce a quantum--wavelet bottleneck that reorganizes latent precipitation features into wavelet sub-bands and performs differentiated structured modulation in the decomposed latent space, aiming to improve the representation of both broad precipitation organization and localized intense structures.
    \item We integrate the refined conditional representation with a rectified-flow-based non-autoregressive decoding strategy for multi-step precipitation nowcasting, with the goal of reducing forecast degradation over later lead times that are particularly relevant to short-term warning.
    \item Under a unified evaluation protocol on the KNMI and SEVIR benchmarks, we find that the proposed framework provides relatively more consistent gains in medium-to-high precipitation regimes, later lead times, and challenging extreme-event cases that are especially relevant to warning-oriented applications.
\end{enumerate}

\section{Related work}

\subsection{Wavelet-based multi-scale representation}

Radar precipitation fields exhibit intertwined multi-scale structures, ranging from broad precipitation organization to localized convective cells with sharp spatial gradients \cite{zheng2025multi}. Preserving such scale diversity is important for representing both the overall organization of precipitation systems and the fine-scale morphology of intense echoes. Although convolutional neural networks can expand receptive fields through hierarchical downsampling, their scale aggregation is often implicit, which may make the preservation of high-frequency details more difficult in challenging nowcasting scenarios \cite{bojinski2023towards}.

The discrete wavelet transform (DWT) provides an explicit decomposition of spatial features into frequency-specific sub-bands \cite{khalili2022image}. In two dimensions, DWT decomposes a feature map into four components, namely LL, LH, HL, and HH. Among them, LL mainly captures coarse-scale background information, whereas LH, HL, and HH retain directional high-frequency structures such as boundaries, local gradients, and abrupt intensity variations. This decomposition is well aligned with radar precipitation fields because it separates broad background information from high-frequency structural details in an interpretable manner.

Wavelet-based representations have been explored in deep models for improving compactness, denoising ability, and robustness. More relevant to radar-related tasks, recent studies suggest that wavelet-driven designs can help preserve sharp boundaries and high-intensity structures. For example, WaveC2R \cite{shi2026wavec2r} introduces a wavelet-driven coarse-to-refined hierarchical framework for radar retrieval and shows that frequency-specific modeling can help decouple large-scale intensity patterns from boundary details on SEVIR-related benchmarks \cite{veillette2020sevir}. These observations suggest that explicit frequency decomposition may also be useful for precipitation nowcasting, where both broad precipitation organization and localized structural details need to be represented.

From a hydrometeorological perspective, explicit frequency separation may be useful because broad background precipitation and localized intense convective structures may play different roles in warning-oriented forecasting. This potential is particularly relevant in hydrological warning because rainfall background extent and localized convective concentration may contribute differently to runoff response and hazard triggering. However, many existing wavelet-based designs mainly emphasize decomposition itself, while its role in conditional generative nowcasting remains less explored.

\subsection{Quantum-inspired representation for structured nonlinear enhancement}

Quantum neural networks and variational quantum circuits (VQCs) \cite{yi2025enhancing} have attracted increasing attention as structured nonlinear transformations for hybrid quantum--classical learning \cite{tabarraei2025variational}. In many recent studies, such modules are used not as replacements for classical deep networks, but as compact nonlinear operators embedded within larger hybrid architectures.

From a representation perspective, quantum-inspired modules can be interpreted as structured nonlinear operators acting on compact latent descriptors. This property is potentially relevant to precipitation nowcasting, where different latent components may correspond to different structural roles across scales. In particular, once precipitation features have been decomposed into wavelet sub-bands, differentiated nonlinear modulation may help enhance components associated with localized boundaries and intensity variations.

In this work, we adopt a quantum-inspired module simulated on classical hardware rather than relying on physical quantum devices \cite{iovane2025quantum}. Its role is not to replace conventional feature extraction, but to provide structured nonlinear modulation after wavelet-based scale disentanglement. Accordingly, the quantum-inspired component in our framework is best understood as a classically simulated sub-band-specific modulation operator rather than as evidence of practical quantum-computing advantage.

\subsection{Generative modeling for precipitation nowcasting}

Generative modeling has become an important direction in precipitation nowcasting because it can better represent forecast uncertainty and often produces sharper precipitation structures than deterministic point-estimation methods. A representative early example is DGMR \cite{chirigati2021accurate}, which showed that deep generative radar nowcasting can achieve skillful and realistic precipitation prediction. More recently, NowcastNet demonstrated strong performance on intense precipitation events by combining physical evolution modeling with deep generative prediction. These studies suggest that generative formulations are well suited to modeling complex spatiotemporal precipitation evolution \cite{ravuri2021skilful}.

Among recent generative paradigms, diffusion-based models have shown strong generation quality in precipitation nowcasting. Diffusion-based methods such as PreDiff \cite{gao2023prediff}, DiffCast \cite{yu2024diffcast}, and related conditional diffusion models \cite{shi2024codicast} further show strong generation quality, although their iterative denoising procedures may still introduce non-negligible inference cost in time-sensitive settings. In operational hydrometeorological warning, such inference cost may become a practical limitation when frequent forecast updates are required.

These developments indicate that the choice of generative formulation can substantially affect forecast sharpness, realism, and computational practicality. At the same time, generation quality also depends on how conditional precipitation information is represented before decoding, which makes the design of the conditional representation particularly relevant. For warning-oriented hydrometeorological forecasting, realistic-looking outputs alone are not sufficient if intense precipitation structures and later lead-time stability are not adequately preserved. Accordingly, a warning-relevant generative nowcasting framework should be judged not only by perceptual realism, but also by its ability to preserve high-impact rainfall structures over operational lead times.

\subsection{Flow-based generative modeling for future-sequence generation}

Flow-based generative formulations provide an alternative by learning a continuous velocity field and generating samples through ODE-based transport. Flow matching \cite{lipman2022flow} and related rectified or straightened flow \cite{dai2025straighten} methods optimize such velocity fields under continuous-time objectives and have recently shown favorable efficiency--fidelity trade-offs in high-dimensional generation tasks. These properties are particularly relevant to precipitation nowcasting, where rapid inference is often desirable in operational settings.

Recent studies have begun to introduce flow-based formulations into meteorological nowcasting. In particular, FlowCast applies conditional flow matching (CFM) \cite{tong2023conditional} to radar nowcasting and shows that a flow-matching objective can produce high-fidelity forecasts with fewer sampling steps than diffusion objectives under similar backbone designs. MeanFlow further proposes a related one-step generative formulation based on average velocity, providing a useful perspective on reducing the number of function evaluations. These studies suggest that flow-based generative modeling is a promising direction for efficient precipitation forecasting.

Our work is related to these methods in that it also adopts a flow-based formulation for non-autoregressive future-sequence generation. The main distinction lies in the conditional representation used by the generative decoder. Instead of relying on a standard latent conditioning pathway, QWRF-Net introduces a wavelet-decomposed and quantum-inspired bottleneck to organize low-frequency background information, high-frequency structural details, and structured nonlinear enhancement before flow-based generation. Therefore, the contribution of the proposed framework lies not only in the use of a flow-based decoder, but also in its integration with a scale-aware conditional representation. This combination is particularly relevant to warning-oriented forecasting, where both structural preservation and later lead-time stability are important.

\section{Methodology}

QWRF-Net is a conditional generative framework for short-term radar-based precipitation nowcasting. As illustrated in Fig.~\ref{fig:overall_arch}, the framework contains two key components: a wavelet-decomposed bottleneck for scale-aware precipitation representation and a flow-based decoder for non-autoregressive future-sequence generation. Given the previous 6 radar frames (30 minutes) at 5-minute intervals, the model generates the next 12 frames (60 minutes).

\subsection{Overall architecture and notation}

QWRF-Net adopts a U-Net-style encoder--decoder backbone to extract multi-scale spatial features from historical radar observations. Skip connections preserve higher-resolution information during up-sampling, while the deepest latent representation is further refined by the proposed wavelet-decomposed bottleneck. For SEVIR, we use VIL as the prediction target, while for KNMI we use the corresponding radar-based precipitation field under the same nowcasting protocol. For both datasets, the target values are normalized to $[0,1]$ during training.

Let $B$ denote the batch size. The historical input sequence and future target sequence are denoted by
\begin{equation}
X_{\mathrm{in}} \in \mathbb{R}^{B \times T_{\mathrm{in}} \times H \times W}, \qquad
X_{\mathrm{tar}} \in \mathbb{R}^{B \times T_{\mathrm{out}} \times H \times W},
\end{equation}
where $T_{\mathrm{in}}=6$, $T_{\mathrm{out}}=12$, and $(H,W)=(288,288)$. Here, $T_{\mathrm{in}}$ and $T_{\mathrm{out}}$ denote the numbers of input and target frames, respectively, while $H$ and $W$ denote the spatial height and width of each frame. For flow-based generation, an initial Gaussian noise sample $Z_0 \sim \mathcal{N}(0,I)$ is drawn with the same shape as $X_{\mathrm{tar}}$. We denote the continuous-time latent state by $Z(t)$ and the learned conditional velocity field by $v_\theta(\cdot)$.

Under this formulation, QWRF-Net parameterizes a time-dependent conditional velocity field
\begin{equation}
v_\theta(Z(t), t, X_{\mathrm{in}}) = \mathrm{QWRF\mbox{-}Net}(Z(t), t, X_{\mathrm{in}}; \theta),
\end{equation}
where $t \in [0,1]$ is embedded using sinusoidal time encoding and injected into the network blocks. The historical sequence $X_{\mathrm{in}}$ provides the condition that guides future-sequence generation. To inject historical context throughout the encoder, multi-scale conditioning features extracted from $X_{\mathrm{in}}$ are fused into the main pathway through a gated fusion operator. This mechanism adaptively balances the main feature stream and the conditioning branch across spatial scales, allowing historical observations to influence both shallow and deep representations. In this way, the historical sequence serves not only as a global condition for generation, but also as guidance for organizing multi-scale latent features before decoding. The resulting deepest latent feature is then processed by the wavelet--quantum bottleneck.

\subsection{Quantum--wavelet bottleneck for scale-aware precipitation representation}

\begin{figure*}[t]
    \centering
    \includegraphics[scale=0.4]{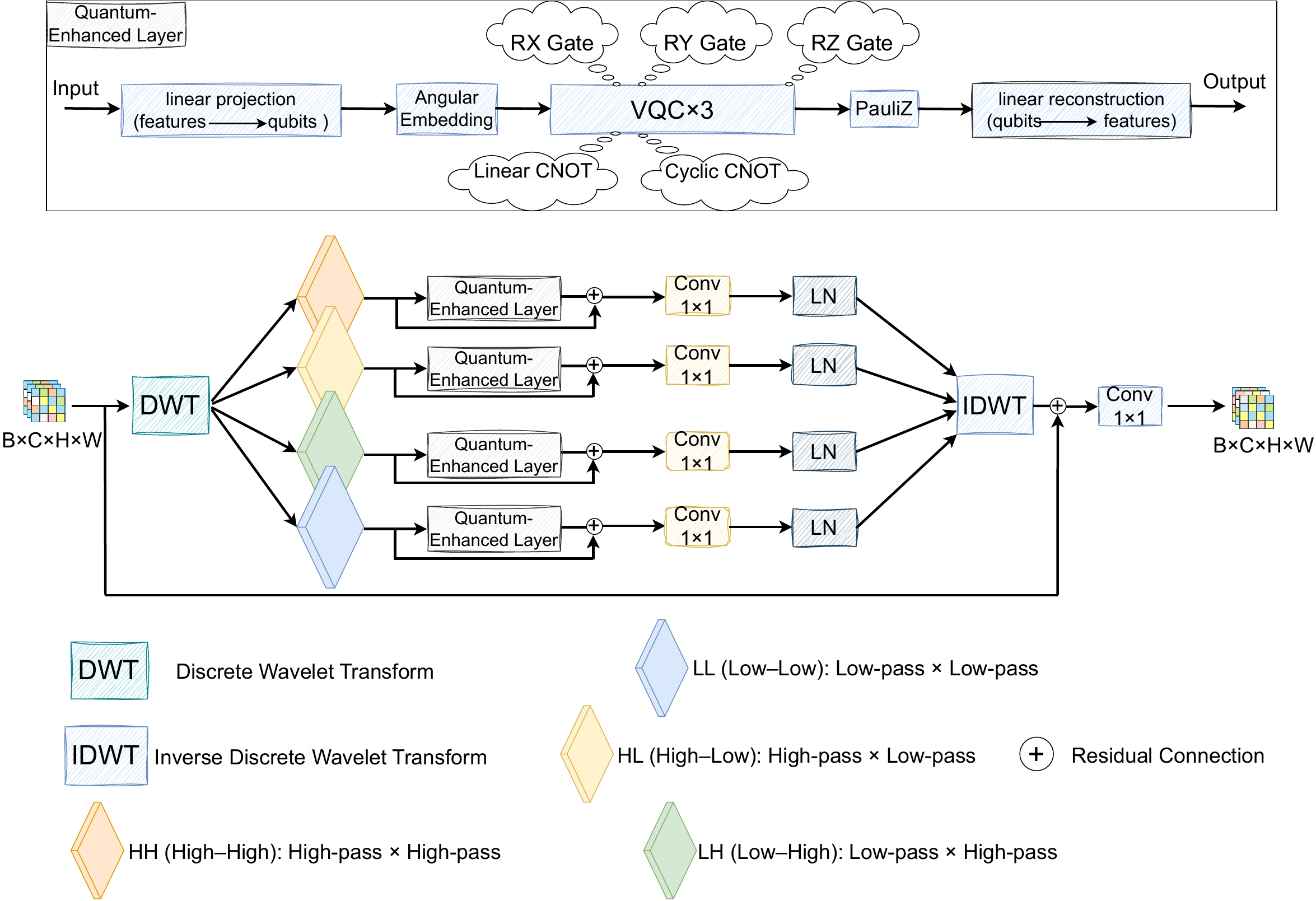}
    \caption{Structure of the quantum--wavelet bottleneck. The bottleneck feature is decomposed into four wavelet sub-bands (LL, LH, HL, and HH) through DWT. Each sub-band is processed by an independent quantum-inspired transformation followed by pointwise projection and normalization. The transformed sub-bands are then fused through IDWT and residual fusion to form the output bottleneck representation.}
    \label{fig:bottleneck}
\end{figure*}

As illustrated in Fig.~\ref{fig:bottleneck}, the quantum--wavelet bottleneck is placed at the deepest level of the U-Net, where the latent representation is relatively compact and therefore suitable for structured transformation. Let the bottleneck feature be
\[
F \in \mathbb{R}^{B \times C \times H_b \times W_b},
\]
where $C$ denotes the channel dimension, and $H_b$ and $W_b$ denote the spatial height and width of the bottleneck feature map, respectively. The bottleneck consists of three stages: wavelet decomposition for scale disentanglement, quantum-inspired sub-band transformation for structured nonlinear enhancement, and reconstruction with residual fusion.

The key idea is to apply structured nonlinear modulation after scale disentanglement rather than before it. Accordingly, the quantum-inspired operator is applied to wavelet sub-bands instead of the original mixed latent feature map, so that low-frequency background information and high-frequency structural details can be modulated in a differentiated manner.

A single-level 2D discrete wavelet transform is first applied to decompose $F$ into four wavelet sub-bands:
\begin{equation}
(F_{LL}, F_{LH}, F_{HL}, F_{HH}) = \mathrm{DWT}(F).
\end{equation}
Here, $F_{LL}$ mainly captures low-frequency background information, whereas $F_{LH}$, $F_{HL}$, and $F_{HH}$ retain directional high-frequency structural details such as boundaries, local gradients, and abrupt intensity variations. This decomposition explicitly reorganizes the latent representation into coarse-scale and fine-scale components before nonlinear modulation. Such decomposition may also be relevant to warning-oriented forecasting because broad precipitation organization and localized intense structures may play different roles in short-term hazard-related prediction. From a hydrological interpretation standpoint, the LL component can be viewed as carrying broad rainfall-background information related to the spatial extent of precipitation, whereas the higher-frequency components are more closely related to intense-core boundaries, localized gradients, and abrupt structural changes. Such explicit separation may be useful when warning relevance depends more strongly on preserving localized convective organization than on reproducing only the average rainfall field.

For each sub-band tensor
\begin{equation}
F_{\mathrm{sub}} \in \{F_{LL},F_{LH},F_{HL},F_{HH}\},
\end{equation}
we apply an independent learnable transformation $\mathcal{Q}(\cdot)$ implemented by a variational quantum circuit simulated on classical hardware. The transformation starts by compressing the spatial feature map into a compact channel descriptor through global average pooling,
\begin{equation}
u_{\mathrm{sub}} = \mathrm{GAP}(F_{\mathrm{sub}}) \in \mathbb{R}^{B \times C},
\end{equation}
so that the sub-band is represented by a low-dimensional summary rather than by the full spatial grid. This design allows the quantum-inspired module to operate as a lightweight structured nonlinear operator on compact sub-band descriptors, rather than replacing spatial feature extraction by the encoder--decoder backbone itself. The descriptor is then projected into a qubit-aligned latent vector,
\begin{equation}
\boldsymbol{\alpha}_{\mathrm{sub}} = W_{\mathrm{enc}} u_{\mathrm{sub}} + b_{\mathrm{enc}} \in \mathbb{R}^{B \times N},
\end{equation}
where $\boldsymbol{\alpha}_{\mathrm{sub}}$ denotes the angle vector used for quantum state encoding, $W_{\mathrm{enc}}$ and $b_{\mathrm{enc}}$ are learnable projection parameters, and $N$ is the number of qubits. The vector $\boldsymbol{\alpha}_{\mathrm{sub}}$ is used as the angular input of the variational quantum circuit.

In this way, the classical sub-band descriptor is encoded into a quantum state by angular embedding,
\begin{equation}
\ket{\psi_{\mathrm{enc}}(\boldsymbol{\alpha}_{\mathrm{sub}})}
=
U_{\mathrm{enc}}(\boldsymbol{\alpha}_{\mathrm{sub}})\ket{0}^{\otimes N},
\end{equation}
where $\ket{0}^{\otimes N}$ denotes the initial all-zero state and $U_{\mathrm{enc}}(\cdot)$ denotes the angle-embedding operator. The encoded state is then processed by a trainable variational circuit,
\begin{equation}
\ket{\psi_{\mathrm{out}}}
=
U(\boldsymbol{\theta})\ket{\psi_{\mathrm{enc}}(\boldsymbol{\alpha}_{\mathrm{sub}})}
=
U(\boldsymbol{\theta})U_{\mathrm{enc}}(\boldsymbol{\alpha}_{\mathrm{sub}})\ket{0}^{\otimes N},
\end{equation}
where $U(\boldsymbol{\theta})$ is parameterized by learnable circuit parameters $\boldsymbol{\theta}$. In practice, this circuit is composed of parameterized single-qubit rotation gates and entangling operations.

To return to the classical feature space, the output quantum state is measured through Pauli-$Z$ expectations,
\begin{equation}
y_i = \langle \psi_{\mathrm{out}} \vert Z_i \vert \psi_{\mathrm{out}} \rangle, \qquad i=1,\dots,N,
\end{equation}
where $Z_i$ denotes the Pauli-$Z$ observable acting on the $i$-th qubit, yielding a classical vector
\begin{equation}
y_{\mathrm{sub}} = [y_1,\dots,y_N] \in \mathbb{R}^{B \times N}.
\end{equation}
This vector is then projected back to the original feature dimension and used as a sub-band-specific modulation term. The overall transformed sub-band is written as
\begin{equation}
F'_{\mathrm{sub}} = \mathcal{T}\!\left(F_{\mathrm{sub}} + \mathcal{Q}(F_{\mathrm{sub}})\right),
\end{equation}
where $\mathcal{Q}(F_{\mathrm{sub}})$ denotes the quantum-inspired modulation reconstructed from $y_{\mathrm{sub}}$, and $\mathcal{T}(\cdot)$ denotes a $1\times1$ convolution followed by layer normalization.

The quantum-inspired transformation can be viewed as a classically simulated structured nonlinear operator that maps each decomposed sub-band descriptor to a sub-band-specific enhancement term. In the proposed framework, its role is to provide differentiated modulation after wavelet-based scale disentanglement rather than to act as an independent predictor. Under this interpretation, the role of the sub-band-specific modulation is to enhance structurally differentiated precipitation information after decomposition rather than to introduce complexity for its own sake. This property may be useful for warning-oriented forecasting, where not all latent precipitation components contribute equally to the identification of intense short-duration rainfall.

This design is particularly suitable after wavelet decomposition because low-frequency background information and high-frequency structural details no longer play identical roles once the bottleneck feature has been reorganized into sub-bands.

As further illustrated in Fig.~\ref{fig:bottleneck}, the transformed wavelet sub-bands are fused back into the spatial domain through the inverse discrete wavelet transform and a residual connection:
\begin{equation}
F_{\mathrm{out}} = \mathrm{Conv}_{1\times1}\!\left(\mathrm{IDWT}(F'_{LL},F'_{LH},F'_{HL},F'_{HH}) + F\right).
\end{equation}
Here, IDWT restores cross-sub-band interactions in the spatial domain, while the residual addition with the original bottleneck feature helps preserve the underlying latent representation during reconstruction. This reconstruction stage also helps maintain the overall spatial continuity of the precipitation field after differentiated sub-band enhancement.

\subsection{Rectified-flow-based future-sequence generation}

After the conditional representation is refined by the wavelet--quantum bottleneck, the future sequence is generated through a rectified-flow-based formulation, so that representation refinement and non-autoregressive generation are connected within the same pipeline. Under this framework, QWRF-Net learns a conditional velocity field that transports an initial Gaussian sample toward the future precipitation sequence.

Let $Z_1 = X_{\mathrm{tar}}$ and sample $Z_0 \sim \mathcal{N}(0,I)$. For a random time $t \sim \mathcal{U}(0,1)$, the interpolation state is defined as
\begin{equation}
Z_t = tZ_1 + (1-t)Z_0.
\end{equation}
This interpolation can be interpreted as a continuous transition from noise to the target precipitation field, providing a natural training pathway for learning temporally coherent precipitation evolution. Such a formulation is also compatible with the view that precipitation evolution is a temporally continuous process, rather than a collection of disconnected frame-wise states.

QWRF-Net is trained to match the corresponding conditional velocity field using
\begin{equation}
\mathcal{L}(\theta) = \mathbb{E}_{Z_0,Z_1,t}\left[\left\|v_\theta(Z_t,t,X_{\mathrm{in}}) - (Z_1-Z_0)\right\|_2^2\right].
\label{eq:flow_loss}
\end{equation}
This objective encourages the model to learn how the noisy latent state should evolve toward the target future sequence under the condition provided by the historical observations. From a forecasting perspective, this learning objective encourages a continuous evolution process rather than a sequence of independently generated steps, which may help preserve temporal stability over later lead times.

At inference time, the future sequence is obtained by solving the conditional ordinary differential equation
\begin{equation}
\frac{dZ}{dt} = v_\theta(Z(t),t,X_{\mathrm{in}}), \qquad Z(0)=Z_0,
\end{equation}
and taking the final state as
\begin{equation}
X_{\mathrm{pred}} = Z(1).
\end{equation}
Unless otherwise stated, we use a forward Euler solver with 50 integration steps for evaluation and visualization. In this way, the model generates the full target sequence jointly rather than through recursive frame-by-frame prediction. This non-autoregressive generation mechanism is expected to reduce error accumulation across time and is therefore particularly relevant to maintaining forecast usefulness over the 60-minute warning-critical horizon considered in this study.

\begin{table*}[t]
\centering
\caption{Quantitative comparison on the KNMI dataset. CSI and HSS are reported at rainfall thresholds of 0.5, 2, 5, 10, and 30 mm/h. MAE, RMSE, and SSIM are also reported. Higher is better for CSI, HSS, and SSIM, while lower is better for MAE and RMSE. The best results are marked in bold, and the second-best results are marked by underlining.}
\label{tab:knmi_results}
\resizebox{\textwidth}{!}{%
\begin{tabular}{lccccccccccccc}
\toprule
\multirow{2}{*}{\textbf{Models}} 
& \multicolumn{5}{c}{\textbf{CSI $\uparrow$}} 
& \multicolumn{5}{c}{\textbf{HSS $\uparrow$}} 
& \multirow{2}{*}{\textbf{MAE}$\downarrow$}
& \multirow{2}{*}{\textbf{RMSE}$\downarrow$}
& \multirow{2}{*}{\textbf{SSIM}$\uparrow$} \\
\cmidrule(lr){2-6} \cmidrule(lr){7-11}
& \textit{r}$\ge$0.5 & \textit{r}$\ge$2 & \textit{r}$\ge$5 & \textit{r}$\ge$10 & \textit{r}$\ge$30
& \textit{r}$\ge$0.5 & \textit{r}$\ge$2 & \textit{r}$\ge$5 & \textit{r}$\ge$10 & \textit{r}$\ge$30
& & & \\
\midrule
ConvLSTM\cite{mihoc2023convsnow}
& 0.6621 & 0.4236 & 0.3458 & 0.1202 & 0.1049 
& 0.4299 & 0.4179 & 0.2416 & 0.1532 & 0.0357
& 6.284 & 7.561 & 0.565 \\

RainNet\cite{ngan2023hybrid}
& 0.6645 & 0.4793 & 0.3129 & 0.1198 & 0.0596 
& 0.5675 & \underline{0.4432} & 0.2345 & 0.0964 & 0.0236
& 7.573 & 9.251 & 0.531 \\

SmaAt-UNet\cite{liao2024short}
& 0.6733 & 0.4689 & 0.3196 & 0.1207 & 0.0698 
& 0.5689 & 0.4369 & 0.2449 & 0.0987 & 0.0567
& 6.254 & 8.652 & 0.545 \\

SimVP\cite{wang2024spatiotemporal}
& 0.6809 & 0.4788 & 0.3094 & 0.1076 & 0.0661 
& 0.5619 & 0.3967 & 0.2578 & 0.1463 & 0.0672
& 6.321 & 8.216 & 0.556 \\

DiffCast\cite{yu2024diffcast}
& 0.6856 & \underline{0.4963} & 0.3196 & \underline{0.1225} & 0.1096 
& 0.5746 & 0.4375 & \underline{0.2586} & 0.1763 & 0.0596
& 6.578 & 7.443 & 0.567 \\

CoDiCast\cite{shi2024codicast}
& \underline{0.6878} & 0.4869 & \underline{0.3596} & 0.1220 & 0.1101 
& 0.5733 & 0.4385 & 0.2536 & 0.1799 & \underline{0.0689}
& 6.491 & 7.125 & 0.571 \\

NowcastNet\cite{zhang2023skilful}
& 0.6864 & 0.4928 & 0.3607 & 0.1232 & \underline{0.1105} 
& \underline{0.5811} & 0.4412 & 0.2581 & \underline{0.1802} & 0.0673
& \underline{5.977} & \underline{7.031} & \underline{0.573} \\

\textbf{QWRF-Net (Ours)} 
& \textbf{0.6963} & \textbf{0.4997} & \textbf{0.3610} & \textbf{0.1236} & \textbf{0.1112} 
& \textbf{0.5869} & \textbf{0.4484} & \textbf{0.2597} & \textbf{0.1812} & \textbf{0.0691}
& \textbf{5.967} & \textbf{7.001} & \textbf{0.585} \\
\bottomrule
\end{tabular}%
}
\end{table*}

\begin{table*}[t]
\centering
\caption{Quantitative comparison on the SEVIR dataset. CSI and HSS are reported on the original VIL intensity scale at thresholds 16, 74, 133, 160, 181, and 219. MAE, RMSE, and SSIM denote the continuous and structural metrics. Higher is better for CSI, HSS, and SSIM, while lower is better for MAE and RMSE. The best results are marked in bold, and the second-best results are marked by underlining.}
\label{tab:sevir_results}
\resizebox{\textwidth}{!}{%
\begin{tabular}{lccccccccccccccc}
\toprule
\multirow{2}{*}{\textbf{Models}} 
& \multicolumn{6}{c}{\textbf{CSI $\uparrow$}} 
& \multicolumn{6}{c}{\textbf{HSS $\uparrow$}} 
& \multirow{2}{*}{\textbf{MAE}$\downarrow$}
& \multirow{2}{*}{\textbf{RMSE}$\downarrow$}
& \multirow{2}{*}{\textbf{SSIM}$\uparrow$} \\
\cmidrule(lr){2-7} \cmidrule(lr){8-13}
& \textit{x}$\ge$16 & \textit{x}$\ge$74 & \textit{x}$\ge$133 & \textit{x}$\ge$160 & \textit{x}$\ge$181 & \textit{x}$\ge$219
& \textit{x}$\ge$16 & \textit{x}$\ge$74 & \textit{x}$\ge$133 & \textit{x}$\ge$160 & \textit{x}$\ge$181 & \textit{x}$\ge$219
& & & \\
\midrule
ConvLSTM\cite{mihoc2023convsnow}
& 0.6191 & 0.5147 & 0.3185 & 0.2361 & 0.2285 & 0.1067
& 0.4151 & 0.3147 & 0.3120 & 0.2211 & 0.1826 & 0.1064
& 8.469 & 9.587 & 0.551 \\

RainNet\cite{ngan2023hybrid}
& 0.5893 & 0.4731 & 0.2991 & 0.2423 & 0.2101 & 0.0699
& 0.3496 & 0.2886 & 0.2112 & 0.1685 & 0.1563 & 0.0721
& 10.312 & 13.542 & 0.523 \\

SmaAt-UNet\cite{liao2024short}
& 0.6375 & 0.5563 & 0.3121 & 0.2651 & 0.2230 & 0.0967
& 0.4508 & 0.4231 & 0.3101 & 0.2257 & 0.1789 & 0.0964
& 9.564 & 11.258 & 0.536 \\

SimVP\cite{wang2024spatiotemporal}
& 0.6496 & 0.5645 & 0.3256 & 0.2756 & 0.2159 & 0.0996
& 0.4501 & 0.4322 & 0.3256 & 0.2311 & 0.1686 & 0.1023
& 8.568 & 9.568 & 0.535 \\

DiffCast\cite{yu2024diffcast}
& 0.6596 & 0.5789 & \underline{0.3696} & 0.2629 & 0.2564 & 0.1302
& \underline{0.4688} & 0.4126 & 0.3316 & \underline{0.2536} & 0.1903 & 0.1033
& 8.481 & 9.621 & 0.549 \\

CoDiCast\cite{shi2024codicast}
& 0.6687 & 0.5796 & 0.3626 & \textbf{0.2789} & \underline{0.2579} & 0.1413
& \textbf{0.4710} & 0.4267 & \underline{0.3466} & 0.2479 & 0.2030 & \underline{0.1120}
& \textbf{8.441} & 9.549 & 0.554 \\

NowcastNet\cite{zhang2023skilful}
& \underline{0.6691} & \underline{0.5861} & 0.3681 & 0.2761 & 0.2571 & \underline{0.1456}
& 0.4661 & \underline{0.4329} & 0.3456 & 0.2531 & \textbf{0.2123} & 0.1118
& 8.463 & \textbf{9.281} & \underline{0.568} \\

\textbf{QWRF-Net (Ours)}
& \textbf{0.6773} & \textbf{0.5869} & \textbf{0.3726} & \underline{0.2763} & \textbf{0.2581} & \textbf{0.1524}
& 0.4612 & \textbf{0.4331} & \textbf{0.3479} & \textbf{0.2579} & \underline{0.2122} & \textbf{0.1131}
& \underline{8.460} & \underline{9.314} & \textbf{0.571} \\
\bottomrule
\end{tabular}%
}
\end{table*}

\begin{table}[t]
\centering
\caption{Performance on the extreme-event subset of SEVIR, defined by peak VIL $\ge 219$ and exceedance ratio of pixels with VIL $>219$ over the 12-frame target stack $\ge 2\%$.}
\label{tab:extreme_results}
\setlength{\tabcolsep}{1pt}
\renewcommand{\arraystretch}{0.88}
\tiny
\resizebox{\linewidth}{!}{%
\begin{tabular}{p{2.15cm}ccc}
\toprule
\textbf{Models} & \textbf{MAE}$\downarrow$ & \textbf{RMSE}$\downarrow$ & \textbf{SSIM}$\uparrow$ \\
\midrule
RainNet\cite{ngan2023hybrid} & 30.241 & 41.215 & 0.103 \\
SmaAt-UNet\cite{liao2024short} & 29.426 & 40.584 & 0.146 \\
SimVP\cite{wang2024spatiotemporal} & 30.544 & 39.337 & 0.169 \\
ConvLSTM\cite{mihoc2023convsnow} & 29.781 & 38.918 & 0.149 \\
DiffCast\cite{yu2024diffcast} & \underline{28.587} & \underline{36.443} & 0.171 \\
CoDiCast\cite{shi2024codicast} & 33.248 & 37.668 & \underline{0.176} \\
NowcastNet\cite{zhang2023skilful} & 29.131 & 36.554 & 0.172 \\
\textbf{QWRF-Net (Ours)} & \textbf{28.054} & \textbf{33.281} & \textbf{0.181} \\
\bottomrule
\end{tabular}%
}
\end{table}

\section{Experiments}

We evaluate QWRF-Net on two public precipitation nowcasting benchmarks, KNMI and SEVIR \cite{veillette2020sevir}, using both categorical and continuous metrics. The experiments are designed to assess not only overall forecasting accuracy, but also structural fidelity, behavior under high-intensity precipitation conditions, and the usefulness of the proposed representation-to-generation design in warning-relevant short-term nowcasting under a unified evaluation setting.

\subsection{Datasets and preprocessing}

We evaluate QWRF-Net on two public precipitation nowcasting datasets, KNMI and SEVIR, which play complementary roles in this study. KNMI provides a regional radar-based nowcasting setting that is closer to practical short-term warning applications, whereas SEVIR offers a larger-sample benchmark for examining structural preservation and robustness under diverse convective conditions, including high-intensity and extreme-event cases. In this sense, the two datasets are used here to assess warning-relevant precipitation nowcasting performance from both a regional radar perspective and a broader benchmark perspective.

The KNMI dataset consists of ground-based radar observations over the Netherlands and surrounding regions, with a temporal resolution of 5 minutes and a spatial resolution of $288 \times 288$. In our experiments, the past 6 frames (30 minutes) are used to predict the next 12 frames (60 minutes). This setting is consistent with the short lead times that are particularly relevant to operational warning-oriented forecasting. To reduce the dominance of non-precipitating samples, sequences without meaningful precipitation signals are filtered out so that the models focus on learning the spatiotemporal evolution of precipitation events that are more relevant to short-term hazard-related applications.

SEVIR is a large-scale storm event imagery dataset for radar and satellite meteorology. We use its vertically integrated liquid (VIL) product with a temporal resolution of 5 minutes. After removing invalid or incomplete samples, 17,321 valid sequences are retained. The original spatial resolution of $384 \times 384$ is resized to $288 \times 288$ for consistency with the KNMI setting and to control computational cost. Under the same forecasting protocol, the past 6 frames are used as input and the subsequent 12 frames are used as prediction targets. In the present work, SEVIR is primarily used to examine model behavior under a standardized large-sample setting and to assess whether the proposed framework remains effective in preserving intense precipitation structures under more challenging convective scenarios.

For both datasets, input and target sequences are normalized to $[0,1]$ during training. For SEVIR, threshold-based categorical scores and extreme-event analysis are additionally reported on the original VIL intensity scale in order to better characterize performance under stronger convective conditions. It should be noted that the use of KNMI and SEVIR in this study is intended to support evaluation of warning-relevant precipitation nowcasting performance, rather than to directly assess downstream hydrological response such as rainfall--runoff, flood routing, or inundation simulation.

\subsection{Implementation details}

QWRF-Net is implemented in PyTorch and uses \texttt{PennyLane} and \texttt{pytorch\_wavelets} for the quantum-inspired transformation and wavelet decomposition, respectively. All experiments are conducted on a multi-GPU server with NVIDIA A6000 GPUs using distributed data parallel (DDP).

Unless otherwise stated, the model takes the past 6 frames as input and predicts the next 12 frames at a spatial resolution of $288 \times 288$. We use the AdamW optimizer with an initial learning rate of $1\times10^{-4}$ and a batch size of 6 per GPU. In the flow-based generation module, the number of sampling steps is set to 10 during training and 50 during evaluation and visualization. The best model checkpoint is selected according to validation loss.

For reproducibility, all compared models are trained and evaluated under the same forecasting protocol, including the same data split, spatial resolution, input--output setting, and preprocessing pipeline. When adapting baseline methods to the unified 6$\rightarrow$12 setting, we keep their overall architectures as close as possible to their original implementations while adjusting only the components necessary for compatibility with the common experimental setup. Model selection is performed using the same validation criterion for all methods.

\begin{figure*}[!t]
    \centering
    \includegraphics[width=\textwidth]{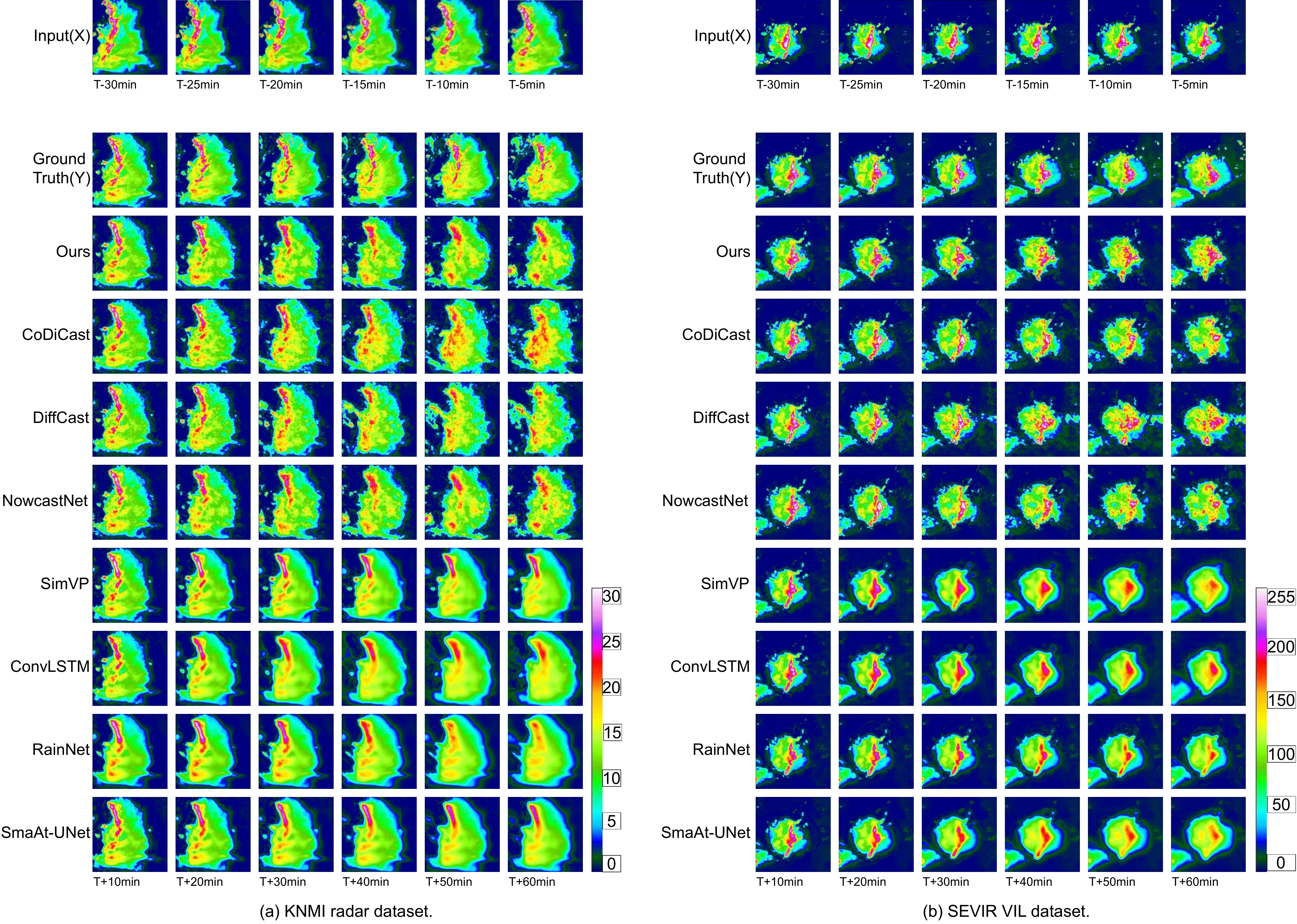}
    \caption{Visual comparison of QWRF-Net with representative methods on two benchmark datasets. (\textbf{a}) KNMI radar dataset. (\textbf{b}) SEVIR VIL dataset. In each panel, the first row shows the input sequence, the second row shows the ground truth, and the remaining rows show the predictions produced by different models from T+10 min to T+60 min.}
    \label{fig:qualitative_comparison}
\end{figure*}

\subsection{Evaluation metrics}

We evaluate model performance using both categorical and continuous metrics. For categorical verification, we report the critical success index (CSI) \cite{mori2012review} and Heidke skill score (HSS) \cite{hyvarinen2014probabilistic}:
\begin{equation}
    \mathrm{CSI} = \frac{\mathrm{TP}}{\mathrm{TP}+\mathrm{FP}+\mathrm{FN}},
    \label{eq:csi}
\end{equation}
\begin{equation}
    \mathrm{HSS} = \frac{2(\mathrm{TP}\cdot \mathrm{TN}-\mathrm{FP}\cdot \mathrm{FN})}{(\mathrm{TP}+\mathrm{FN})(\mathrm{FN}+\mathrm{TN})+(\mathrm{TP}+\mathrm{FP})(\mathrm{FP}+\mathrm{TN})},
    \label{eq:hss}
\end{equation}
where TP, FP, FN, and TN denote true positives, false positives, false negatives, and true negatives, respectively. In warning-oriented forecasting, CSI is particularly relevant because it reflects successful detection of threshold-exceeding precipitation events, while HSS provides a skill-based assessment beyond chance agreement.

To complement threshold-based evaluation, we additionally report mean absolute error (MAE) \cite{error2016mean}, root mean squared error (RMSE) \cite{hodson2022root}, and structural similarity index measure (SSIM) \cite{nilsson2020understanding}, which quantify intensity accuracy and structural consistency. Given the predicted sequence $X_{\mathrm{pred}}$ and the ground-truth sequence $X_{\mathrm{tar}}$, MAE and RMSE are defined as
\begin{equation}
    \mathrm{MAE} = \frac{1}{N}\sum_{i=1}^{N} \left| X_{\mathrm{pred}}^{(i)} - X_{\mathrm{tar}}^{(i)} \right|,
    \label{eq:mae}
\end{equation}
\begin{equation}
    \mathrm{RMSE} = \sqrt{\frac{1}{N}\sum_{i=1}^{N} \left( X_{\mathrm{pred}}^{(i)} - X_{\mathrm{tar}}^{(i)} \right)^2},
    \label{eq:rmse}
\end{equation}
where $N$ denotes the total number of evaluated pixels over all forecast frames. MAE and RMSE quantify intensity errors that may be relevant when considering the potential downstream use of precipitation nowcasts in rainfall-driven hydrological applications.

For structural similarity, we adopt SSIM, defined as
\begin{equation}
    \mathrm{SSIM}(x,y)=\frac{(2\mu_x\mu_y + C_1)(2\sigma_{xy}+C_2)}{(\mu_x^2+\mu_y^2+C_1)(\sigma_x^2+\sigma_y^2+C_2)},
    \label{eq:ssim}
\end{equation}
where $\mu_x$ and $\mu_y$ are the mean intensities of the prediction and ground truth, $\sigma_x^2$ and $\sigma_y^2$ are their variances, $\sigma_{xy}$ is the covariance, and $C_1$, $C_2$ are stabilization constants. For warning-relevant precipitation forecasting, SSIM is particularly useful because preserving intense core location and boundary structure may matter more than mean intensity agreement alone.

For KNMI, CSI and HSS are computed at rain-rate thresholds of 0.5, 2, 5, 10, and 30 mm/h, following common evaluation settings. For SEVIR, threshold-based scores are computed on the original VIL intensity scale using thresholds $\{16, 74, 133, 160, 181, 219\}$, covering precipitation regimes from weak echoes to extreme convective intensity.

To further assess model behavior under high-impact precipitation, we define an extreme-event subset on SEVIR. A sample is regarded as extreme if, over the 12-frame target stack, the peak VIL intensity is at least 219 and the exceedance ratio of pixels with VIL $>219$ is no less than 2\%.

\begin{figure*}[!t]
  \centering
  \includegraphics[width=\linewidth]{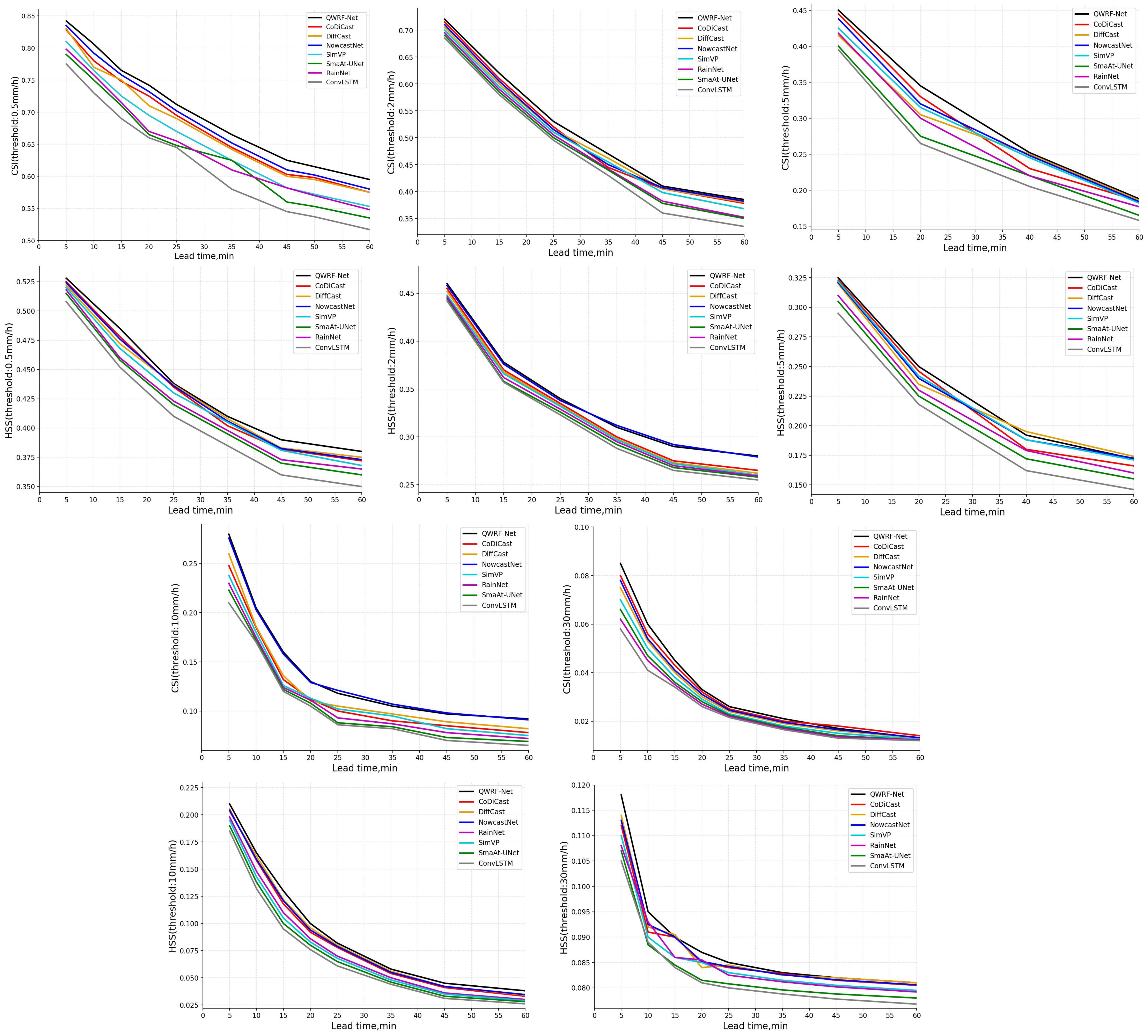}
  \caption{Lead-time curves of CSI and HSS for different models on the KNMI dataset.}
  \label{fig:curves}
\end{figure*}

\subsection{Baselines and fair comparison protocol}

We compare QWRF-Net against representative discriminative baselines, including RainNet, SmaAt-UNet, SimVP, and ConvLSTM, as well as strong generative baselines, including DiffCast, CoDiCast, and NowcastNet. These baselines were selected to represent several commonly used families of data-driven nowcasting models, including recurrent, convolutional encoder--decoder, and recent generative approaches.

To improve comparability, all models are trained and evaluated under the same data split, spatial resolution, input--output setting (6$\rightarrow$12), and preprocessing pipeline. For baseline methods whose original implementations were not designed exactly for this setting, we adapt them to the unified protocol while preserving their main architectural characteristics. Hyperparameters are adjusted within a comparable training setting, and the best checkpoint of each method is selected using the same validation criterion.

We note that the compared generative baselines differ in their original optimization and sampling configurations. Therefore, our goal is not to reproduce every method under its task-specific best-case setting, but to assess their behavior under a common and controlled nowcasting protocol. Under this setting, all methods are compared using the same training/validation split, input and prediction horizon, and evaluation pipeline, so that the relative differences are more directly attributable to model design rather than to mismatched experimental conditions. The purpose of this comparison is not only to compare predictive accuracy across model families, but also to examine which design choices appear more favorable for preserving warning-relevant precipitation structures under a common short-term nowcasting setting.

\begin{table*}[t]
\centering
\caption{Ablation results on the KNMI dataset. CSI and HSS are reported at rainfall thresholds of 0.5, 2, 5, 10, and 30 mm/h. MAE, RMSE, and SSIM denote the continuous and structural metrics. Higher is better for CSI, HSS, and SSIM, while lower is better for MAE and RMSE. The best results are marked in bold.}
\label{tab:knmi_ablation}
\resizebox{\textwidth}{!}{%
\begin{tabular}{lccccccccccccc}
\toprule
\multirow{2}{*}{\textbf{Models}}
& \multicolumn{5}{c}{\textbf{CSI $\uparrow$}}
& \multicolumn{5}{c}{\textbf{HSS $\uparrow$}}
& \multirow{2}{*}{\textbf{MAE}$\downarrow$}
& \multirow{2}{*}{\textbf{RMSE}$\downarrow$}
& \multirow{2}{*}{\textbf{SSIM}$\uparrow$} \\
\cmidrule(lr){2-6} \cmidrule(lr){7-11}
& \textit{r}$\ge$0.5 & \textit{r}$\ge$2 & \textit{r}$\ge$5 & \textit{r}$\ge$10 & \textit{r}$\ge$30
& \textit{r}$\ge$0.5 & \textit{r}$\ge$2 & \textit{r}$\ge$5 & \textit{r}$\ge$10 & \textit{r}$\ge$30
& & & \\
\midrule
QW-Net 
& 0.6812 & 0.4421 & 0.2933 & 0.0822 & 0.0651 
& 0.4355 & 0.3622 & 0.2215 & 0.1133 & 0.0515
& 6.812 & 8.236 & 0.548 \\

QWRF-Net-A 
& 0.6581 & 0.4533 & 0.3011 & 0.0911 & 0.0722 
& 0.4422 & 0.3711 & 0.2311 & 0.1211 & 0.0311
& 6.954 & 8.487 & 0.541 \\

QWRF-Net-B 
& 0.6811 & 0.4801 & 0.3301 & 0.1011 & 0.0815 
& 0.5701 & 0.4211 & 0.2401 & 0.1501 & 0.0511
& 6.421 & 7.884 & 0.559 \\

QWRF-Net-C 
& 0.6755 & 0.4855 & 0.3355 & 0.1055 & 0.0911 
& 0.5655 & 0.4255 & 0.2455 & 0.1555 & 0.0601
& 6.318 & 7.701 & 0.563 \\

QWRF-Net-D 
& 0.6701 & 0.4888 & 0.3411 & 0.1151 & 0.1051 
& 0.5601 & 0.4301 & 0.2501 & 0.1701 & 0.0651
& 6.184 & 7.502 & 0.569 \\

QWRF-Net-E 
& 0.6901 & 0.4955 & 0.3501 & 0.1101 & 0.1001 
& 0.5801 & 0.4401 & 0.2555 & 0.1601 & 0.0611
& 6.072 & 7.211 & 0.576 \\

\textbf{QWRF-Net (Ours)} 
& \textbf{0.6963} & \textbf{0.4997} & \textbf{0.3610} & \textbf{0.1236} & \textbf{0.1112}
& \textbf{0.5869} & \textbf{0.4484} & \textbf{0.2597} & \textbf{0.1812} & \textbf{0.0691}
& \textbf{5.967} & \textbf{7.001} & \textbf{0.585} \\
\bottomrule
\end{tabular}%
}
\end{table*}

\begin{table*}[t]
\centering
\caption{Ablation results on the SEVIR dataset. CSI and HSS are reported on the original VIL intensity scale at thresholds 16, 74, 133, 160, 181, and 219. MAE, RMSE, and SSIM denote the continuous and structural metrics. Higher is better for CSI, HSS, and SSIM, while lower is better for MAE and RMSE. The best results are marked in bold.}
\label{tab:sevir_ablation}
\resizebox{\textwidth}{!}{%
\begin{tabular}{lccccccccccccccc}
\toprule
\multirow{2}{*}{\textbf{Models}}
& \multicolumn{6}{c}{\textbf{CSI $\uparrow$}}
& \multicolumn{6}{c}{\textbf{HSS $\uparrow$}}
& \multirow{2}{*}{\textbf{MAE}$\downarrow$}
& \multirow{2}{*}{\textbf{RMSE}$\downarrow$}
& \multirow{2}{*}{\textbf{SSIM}$\uparrow$} \\
\cmidrule(lr){2-7} \cmidrule(lr){8-13}
& \textit{x}$\ge$16 & \textit{x}$\ge$74 & \textit{x}$\ge$133 & \textit{x}$\ge$160 & \textit{x}$\ge$181 & \textit{x}$\ge$219
& \textit{x}$\ge$16 & \textit{x}$\ge$74 & \textit{x}$\ge$133 & \textit{x}$\ge$160 & \textit{x}$\ge$181 & \textit{x}$\ge$219
& & & \\
\midrule
QW-Net 
& 0.6768 & 0.5122 & 0.3155 & 0.2411 & 0.2133 & 0.0822 
& 0.3422 & 0.2933 & 0.2244 & 0.1733 & 0.1544 & 0.1094
& 8.972 & 10.441 & 0.529 \\

QWRF-Net-A 
& 0.6322 & 0.5211 & 0.3211 & 0.2488 & 0.2211 & 0.0988 
& 0.3611 & 0.3011 & 0.2311 & 0.1811 & 0.1622 & 0.0811
& 9.154 & 10.862 & 0.522 \\

QWRF-Net-B 
& 0.6611 & 0.5701 & 0.3501 & 0.2601 & 0.2401 & 0.1201 
& 0.4501 & 0.4101 & 0.3201 & 0.2301 & 0.1901 & 0.1001
& 8.714 & 9.944 & 0.548 \\

QWRF-Net-C 
& 0.6555 & 0.5755 & 0.3555 & 0.2655 & 0.2455 & 0.1301 
& 0.4533 & 0.4155 & 0.3301 & 0.2401 & 0.1955 & 0.1051
& 8.603 & 9.778 & 0.553 \\

QWRF-Net-D 
& 0.6501 & 0.5788 & 0.3601 & 0.2701 & 0.2501 & 0.1451 
& 0.4488 & 0.4201 & 0.3355 & 0.2501 & 0.2051 & 0.1101
& 8.521 & 9.602 & 0.561 \\

QWRF-Net-E 
& 0.6701 & 0.5801 & 0.3655 & 0.2688 & 0.2551 & 0.1401 
& 0.4555 & 0.4255 & 0.3401 & 0.2455 & 0.2001 & 0.1088
& 8.493 & 9.497 & 0.564 \\

\textbf{QWRF-Net (Ours)} 
& \textbf{0.6773} & \textbf{0.5869} & \textbf{0.3726} & \textbf{0.2763} & \textbf{0.2581} & \textbf{0.1524}
& \textbf{0.4612} & \textbf{0.4331} & \textbf{0.3479} & \textbf{0.2579} & \textbf{0.2122} & \textbf{0.1131}
& \textbf{8.460} & \textbf{9.314} & \textbf{0.571} \\
\bottomrule
\end{tabular}%
}
\end{table*}

\subsection{Quantitative results}

We first report quantitative results from both categorical verification and continuous/structural assessment so as to examine detection performance, intensity accuracy, and morphological consistency under a unified setting.

Table~\ref{tab:knmi_results} reports the categorical and continuous/structural results on KNMI. QWRF-Net achieves the strongest overall results under the present setting, including the highest CSI and HSS values across the reported thresholds, together with the lowest MAE and RMSE and the highest SSIM. Although the margins over the strongest baselines are moderate for some metrics, the gains are relatively consistent across threshold-based verification and reconstruction-oriented evaluation.

From a hydrometeorological forecasting perspective, the advantage of QWRF-Net becomes more visible at medium and high rain-rate thresholds, where prediction is generally more difficult and where preserving strong precipitation cores is particularly important for warning-relevant nowcasting. The favorable SSIM, together with the improvements at higher thresholds, further suggests that the proposed representation-to-generation design helps preserve structural details over the forecast horizon. Lower intensity errors and improved structural consistency may provide a more useful precipitation basis for downstream hydrological use than less stable nowcasts. For warning-oriented use, the improvements at $r\ge10$ and $r\ge30$ are particularly noteworthy because these thresholds are more closely associated with high-impact short-duration rainfall than low-threshold background precipitation alone. In this sense, the gains of QWRF-Net are not limited to average reconstruction quality, but also extend to rainfall regimes that are more relevant to flood-triggering conditions.

Table~\ref{tab:sevir_results} reports the categorical and continuous/structural results on SEVIR. QWRF-Net achieves favorable performance at medium-to-high VIL thresholds and attains the highest SSIM, while remaining competitive in MAE and RMSE. Compared with strong baselines such as NowcastNet and CoDiCast, the gains are generally moderate in absolute magnitude, but they are more consistent at higher intensity thresholds and in structural-fidelity-related evaluation.

This observation is important because average performance alone may obscure model behavior under stronger convective activity. Under the present setting, the advantage of QWRF-Net is more evident in regimes where preserving intense structures and maintaining boundary sharpness become increasingly difficult. Such behavior is particularly relevant to short-term hazard-oriented forecasting, where the location, continuity, and intensity of convective precipitation cores may matter more than average reconstruction quality alone. Therefore, the SEVIR results support the view that the proposed framework is potentially useful for maintaining structural consistency under challenging precipitation conditions rather than only improving easier or low-intensity cases. Such behavior may be particularly relevant when precipitation nowcasts are considered as potential upstream inputs for warning-related or hydrological applications that are sensitive to the location and continuity of intense rainfall structures.

Table~\ref{tab:extreme_results} reports MAE, RMSE, and SSIM on the extreme-event subset of SEVIR. Since average metrics over the full test set may mask model behavior under high-impact precipitation, this subset provides a more focused evaluation of challenging convective cases with strong intensity and sufficient spatial extent. On this subset, QWRF-Net achieves the lowest MAE and RMSE together with the highest SSIM. These results suggest that the proposed framework remains relatively robust when the prediction target contains more intense precipitation cores and sharper structural variations, which is particularly relevant to warning-oriented nowcasting. This behavior is important because extreme-event cases are often the most consequential for flash-flood and urban inundation warning, while also being the cases in which structural distortion is most damaging to forecast usefulness. The improved robustness of QWRF-Net on this subset therefore further suggests its potential relevance for high-impact short-term warning and related downstream use.

\subsection{Practical relevance for warning-oriented nowcasting}

Although this study focuses on precipitation nowcasting rather than downstream hydrological simulation, the observed improvements are potentially relevant to hydrometeorological early warning. In particular, gains at medium-to-high precipitation thresholds, better preservation of intense precipitation cores, and relatively slower degradation at later lead times are all desirable properties in warning-oriented applications. These characteristics may provide a more useful precipitation basis for subsequent flood-related analysis and hydrological modeling. In particular, structure-preserving nowcasts may be more suitable as precipitation inputs for distributed hydrological or inundation-related models than forecasts that achieve similar average error but lose intense-core organization.

\begin{figure*}[t]
    \centering
    \includegraphics[width=\textwidth]{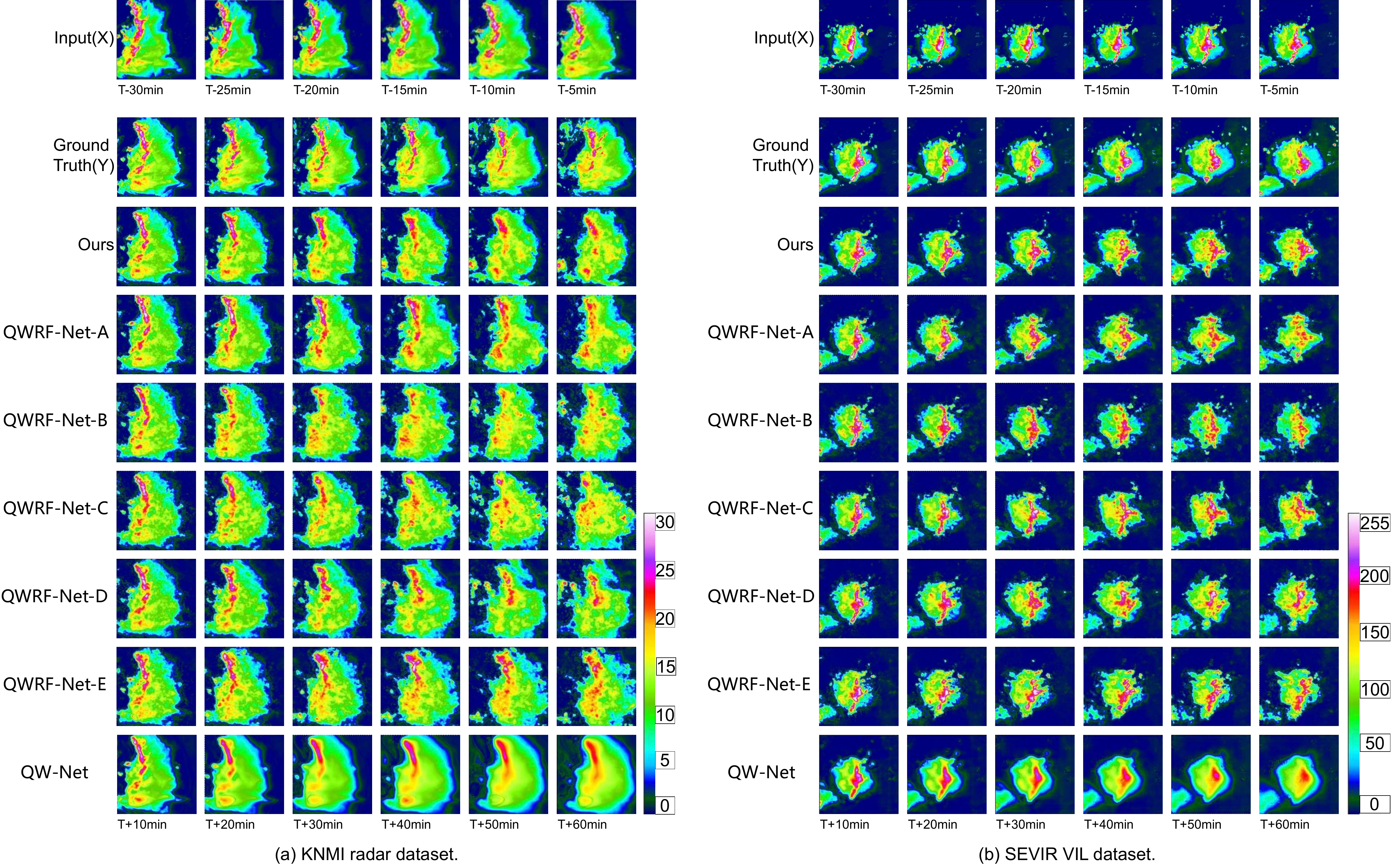}
    \caption{Visual comparison of QWRF-Net and its ablation variants on two benchmark datasets. (\textbf{a}) KNMI radar dataset. (\textbf{b}) SEVIR VIL dataset. In each panel, the first row shows the input sequence, the second row shows the ground truth, and the remaining rows show the predictions of the full model and different ablation variants from T+10 min to T+60 min.}
    \label{fig:ablation_visual}
\end{figure*}

\subsection{Qualitative results}

To provide a compact visual comparison, the qualitative results on KNMI and SEVIR are organized into a unified figure, where the two datasets are shown as separate panels for direct comparison. Representative cases from the two benchmark datasets are shown in Fig.~\ref{fig:qualitative_comparison}. As shown in Fig.~\ref{fig:qualitative_comparison}(a), corresponding to the KNMI radar dataset, QWRF-Net better preserves the morphology and intensity distribution of the main precipitation body, especially at middle and later lead times. As shown in Fig.~\ref{fig:qualitative_comparison}(b), corresponding to the SEVIR VIL dataset, QWRF-Net produces predictions that are visually more consistent with the ground truth in core location, boundary sharpness, and structural continuity.

Compared with representative discriminative baselines and strong generative baselines such as DiffCast, CoDiCast, and NowcastNet, the proposed method shows clearer advantages in maintaining intense precipitation cores and fine-scale structures in the illustrated cases. These qualitative observations are consistent with the quantitative improvements at medium-to-high thresholds and further suggest that the proposed framework may be useful for preserving precipitation structures that are relevant to short-term warning. From a hydrometeorological perspective, better preservation of core location and boundary sharpness may be particularly important where localized intense rainfall governs short-term warning relevance and may also be relevant to the potential downstream use of such nowcasts.

\subsection{Lead-time analysis}

Fig.~\ref{fig:curves} presents the lead-time curves of CSI and HSS on the KNMI dataset. As the forecast horizon increases, QWRF-Net maintains strong performance and exhibits a relatively slower degradation trend than the compared baselines. This behavior is broadly consistent with the design motivation of the non-autoregressive flow-based generation mechanism.

From an operational perspective, stable behavior across the full forecast horizon is often more valuable than isolated gains at a single lead time. In practical warning settings, forecast usefulness depends not only on early lead-time accuracy but also on maintaining reliable information throughout the operational forecast window. In particular, the 40--60 minute range is often critical for short-term preparedness and is also where many models show substantial structural degradation. Overall, Fig.~\ref{fig:curves} provides supportive evidence that the proposed framework can maintain relatively stable forecast quality at later lead times under the current evaluation setting. This later lead-time stability is especially relevant in practice, because forecast usefulness for warning support often depends more on whether quality remains acceptable across the full lead-time window than on isolated gains at the earliest frames.

\section{Ablation study}

To further examine the effectiveness of the main components in QWRF-Net, we conduct ablation experiments on both KNMI and SEVIR. The purpose is not only to test whether each component contributes to the final performance, but also to examine whether the proposed design is beneficial for preserving structurally important precipitation information before multi-step generation.

\subsection{Ablation setup}

We construct the following ablation variants:

\begin{itemize}
    \item \textbf{QWRF-Net-A (Simple Bottleneck -- No Wavelet, No Quantum):}
    This variant retains the overall U-Net-style architecture and the rectified-flow generation mechanism, but replaces the quantum--wavelet bottleneck with a standard convolutional bottleneck. It is used to assess the contribution of the proposed bottleneck design as a whole.

    \item \textbf{QWRF-Net-B (Wavelet-Only -- No Quantum):}
    This variant preserves the DWT-based decomposition and reconstruction, but replaces the quantum-inspired transformation with standard convolutional sub-band processing. It is used to isolate the role of explicit wavelet decomposition.

    \item \textbf{QWRF-Net-C (Wavelet + Classical MLP):}
    This variant preserves the wavelet decomposition, but replaces the quantum-inspired transformation with classical MLP-based mappings. It is used to examine whether differentiated sub-band modulation provides benefits beyond conventional nonlinear mixing.

    \item \textbf{QWRF-Net-D (Shared Quantum Processor):}
    This variant uses both DWT and quantum-inspired processing, but all four sub-bands share the same quantum processor rather than using independent processors. It is used to test the necessity of sub-band-specific processing after decomposition.

    \item \textbf{QWRF-Net-E (Quantum-Only -- No Wavelet):}
    This variant removes DWT/IDWT and applies the quantum-inspired transformation directly to the bottleneck feature map. It is used to evaluate whether nonlinear enhancement without explicit scale disentanglement is sufficient.

    \item \textbf{QW-Net (Without Rectified Flow):}
    This variant retains the quantum--wavelet bottleneck but replaces the rectified-flow generation mechanism with a conventional discriminative prediction head. It is used to evaluate the contribution of the flow-based non-autoregressive generation strategy.
\end{itemize}

In addition to CSI and HSS, we also report MAE, RMSE, and SSIM in the ablation study so as to examine whether the observed differences among variants are consistent in terms of intensity accuracy and structural fidelity.

\subsection{Analysis of ablation results}

The ablation results in Tables~\ref{tab:knmi_ablation} and~\ref{tab:sevir_ablation} lead to several observations.

First, the complete QWRF-Net achieves the best overall results on both datasets, particularly at medium-to-high precipitation thresholds. This suggests that the full combination of wavelet-based scale disentanglement, differentiated sub-band modulation, and rectified-flow-based future-sequence generation is more effective than the corresponding partial variants under the present setting. From a warning-oriented perspective, this is especially important because improvements at higher thresholds are more relevant to intense precipitation regimes.

Second, replacing the quantum--wavelet bottleneck with a standard convolutional bottleneck (QWRF-Net-A) leads to a clear performance drop on both datasets, indicating that the proposed bottleneck contributes meaningfully to the quality of the conditional representation before generative decoding.

Third, the comparisons among QWRF-Net-B, QWRF-Net-C, QWRF-Net-D, and QWRF-Net-E help rule out several simpler explanations. QWRF-Net-B shows that explicit wavelet decomposition alone is beneficial, but not sufficient to match the full model. QWRF-Net-C indicates that the gain is not merely due to inserting a generic nonlinear mapping after decomposition. QWRF-Net-D shows that using a shared processor across all sub-bands is less effective than using sub-band-specific modulation, suggesting that the decomposed components indeed play different roles and are better handled in a differentiated manner. The comparison between QWRF-Net-E and the full model further suggests that nonlinear enhancement is more effective after wavelet-based scale disentanglement than when it is directly imposed on the original mixed bottleneck feature.

Taken together, these results support a more specific interpretation of the proposed bottleneck design: the observed benefit does not arise simply from adding wavelet decomposition, inserting a generic nonlinear operator, or increasing architectural complexity. Rather, the improvement is associated with the intended sequence of operations, namely decomposition first and structured sub-band-specific modulation afterward. This suggests that explicit organization of precipitation information by scale may be an important prerequisite for enhancing warning-relevant structural features within the present framework.

Fourth, the comparison with QW-Net highlights the role of the rectified-flow generation mechanism. QW-Net remains competitive on some lower-threshold metrics, but the advantage of the full QWRF-Net becomes more visible at higher thresholds, especially on SEVIR at \textit{x}$\ge$219 and on KNMI at \textit{r}$\ge$10 and \textit{r}$\ge$30. This trend is consistent with the role of flow-based generation in maintaining forecast quality under more challenging precipitation regimes and later lead times.

The visual comparisons in Fig.~\ref{fig:ablation_visual} are broadly consistent with the quantitative ablation results. As illustrated in Fig.~\ref{fig:ablation_visual}(a) and Fig.~\ref{fig:ablation_visual}(b), on both KNMI and SEVIR, the predictions of the complete QWRF-Net remain closer to the ground truth in morphology, intensity distribution, and spatial continuity of strong precipitation cores, particularly at middle and late forecast stages. In contrast, several ablation variants exhibit more noticeable smoothing, structural distortion, or loss of local detail, which further supports the effectiveness of the full design.

Overall, the ablation results support the design logic of QWRF-Net, namely that differentiated sub-band modulation is more effective when applied after wavelet-based scale disentanglement within the present framework. This finding is relevant because preserving intense, structured precipitation information is especially important in warning-oriented short-term nowcasting.

\section{Conclusion}

In this work, we presented QWRF-Net, a quantum--wavelet framework with rectified flow for short-term precipitation nowcasting. The framework is designed to address two closely related challenges in nowcasting: representing intertwined multi-scale precipitation structures and reducing degradation in multi-step future prediction. To this end, QWRF-Net combines wavelet-based latent decomposition, quantum-inspired sub-band modulation, and rectified-flow-based future-sequence generation within a unified framework.

Experimental results on KNMI and SEVIR show that QWRF-Net provides favorable overall performance under the unified 6$\rightarrow$12 setting. In addition to improving average predictive quality, the proposed model shows relatively more consistent gains in structurally challenging regimes, including medium-to-high precipitation thresholds, later lead times, and an extreme-event subset. The ablation results are broadly consistent with the intended design logic, indicating that wavelet-based decomposition, differentiated sub-band modulation, and flow-based generation provide complementary benefits when organized in the proposed sequence.

From a hydrometeorological perspective, these findings suggest that jointly improving conditional precipitation representation and future-sequence generation is a useful direction for warning-relevant short-term precipitation forecasting, especially when preserving intense precipitation structures and later lead-time stability is important. These improvements may also provide a more useful precipitation basis for subsequent warning-related and hydrological applications.

Several limitations should also be noted. The present study focuses on a 60-minute forecasting horizon, and the conclusions should therefore be interpreted within this setting. In addition, the quantum-inspired module is simulated on classical hardware and should be understood as a structured nonlinear operator rather than as evidence of practical quantum-computing advantage. The current bottleneck design also does not yet model richer hierarchical cross-scale interactions, and the study does not directly assess how any improvement in nowcast quality may translate into downstream hydrological response.

Future work will examine whether the proposed framework remains effective under longer forecast horizons, higher spatial resolutions, and more diverse regional precipitation datasets. It will also investigate the coupling of the predicted precipitation with distributed hydrological or inundation-related models to assess its practical value for warning-oriented decision support more directly. In this respect, QWRF-Net may serve as a potentially useful link between radar-based precipitation nowcasting and downstream hydrological or warning-related applications.

\section*{Acknowledgments}

The authors would like to thank the Royal Netherlands Meteorological Institute (KNMI) for providing the radar dataset used in this study. The authors also acknowledge the providers of the SEVIR dataset, including MIT Lincoln Laboratory, for making the storm event imagery publicly available. The authors sincerely thank Yibin University and the Yibin Municipal Science and Technology Bureau for their support of this research. This work was supported in part by the Major Project of Yibin University under Grant 2025XJZD01, and in part by the Science and Technology Program of the Yibin Municipal Science and Technology Bureau under Grant 2025JC008.

\bibliographystyle{elsarticle-num}
\bibliography{main}

\end{document}